\documentclass[runningheads]{llncs}
\usepackage{graphicx}
\usepackage{authblk}
\usepackage{graphicx}
\usepackage{subfigure}
\usepackage{array}
\usepackage{float}
\usepackage{booktabs}
\usepackage{bbding}
\usepackage{etoolbox}
\usepackage{color}
\usepackage{soul}
\usepackage{algorithm}
\usepackage{algorithmic}
\usepackage{cite}
\usepackage{balance}
\usepackage{amsmath, amssymb}
\usepackage{makecell}
\usepackage{multirow}
\usepackage{hyperref}
\begin{document}
\title{Revisiting Distillation for Continual Learning on Visual Question Localized-Answering in Robotic Surgery}
\titlerunning{Revisiting Distillation for Continual VQLA in Robotic Surgery}
%

%
\authorrunning{L. Bai et al.}
\author{Long Bai\inst{1~\star} 
\and Mobarakol Islam\inst{2} 
\thanks{Long Bai and Mobarakol Islam are co-first authors.} 
\and Hongliang Ren\inst{1,3} 
\thanks{Corresponding author.}}
\institute{Dept. of Electronic Engineering, The Chinese University of Hong Kong (CUHK), Hong Kong SAR, China
\and Wellcome/EPSRC Centre for Interventional and Surgical Sciences (WEISS), University College London, London, UK
\and Shun Hing Institute of Advanced Engineering, CUHK, Hong Kong SAR, China\\
\email{b.long@link.cuhk.edu.hk, mobarakol.islam@ucl.ac.uk, hlren@ee.cuhk.edu.hk}}
\maketitle              
\begin{abstract}
The visual-question localized-answering (VQLA) system can serve as a knowledgeable assistant in surgical education. Except for providing text-based answers, the VQLA system can highlight the interested region for better surgical scene understanding. However, deep neural networks (DNNs) suffer from catastrophic forgetting when learning new knowledge. Specifically, when DNNs learn on incremental classes or tasks, their performance on old tasks drops dramatically. Furthermore, due to medical data privacy and licensing issues, it is often difficult to access old data when updating continual learning (CL) models. Therefore, we develop a non-exemplar continual surgical VQLA framework, to explore and balance the rigidity-plasticity trade-off of DNNs in a sequential learning paradigm. We revisit the distillation loss in CL tasks, and propose rigidity-plasticity-aware distillation (RP-Dist) and self-calibrated heterogeneous distillation (SH-Dist) to preserve the old knowledge. The weight aligning (WA) technique is also integrated to adjust the weight bias between old and new tasks. We further establish a CL framework on three public surgical datasets in the context of surgical settings that consist of overlapping classes between old and new surgical VQLA tasks. With extensive experiments, we demonstrate that our proposed method excellently reconciles learning and forgetting on the continual surgical VQLA over conventional CL methods. Our code is publicly accessible at \href{https://github.com/longbai1006/CS-VQLA}{github.com/longbai1006/CS-VQLA}.

\end{abstract}
\section{Introduction}
Trustworthy and reliable visual question-answering (VQA) models have proved their potential in the medical domain~\cite{lin2021medical,seenivasan2022surgical}. As medical deep learning (DL) develops exuberantly~\cite{bai2021influence,che2023image}, a DL-based surgical VQA system~\cite{seenivasan2022surgical} has been developed as a surgical training and popularization tool for junior surgeons, medical students, and patients. However, one pivotal problem with surgical VQA is the lack of localized answers. VQA can provide the answer to the question, but cannot relate the answers to its localization at an instance level. Surgical scenarios with various similar instruments and actions may further confuse the learners. Answers with localization can further assist learners in dealing with confusion. In this case, a surgical visual-question localized-answering (VQLA) system can thereby be established for effective surgical training and scene understanding~\cite{bai2023surgical}.

Meanwhile, catastrophic forgetting has become a largely discussed topic in deep neural networks. Deep neural networks (DNNs) shall abruptly and drastically forget old knowledge when learning new~\cite{li2017lwf}. Various continual learning (CL) methods have been proposed to mitigate catastrophic forgetting and study the balance of rigidity and plasticity in deep models~\cite{li2017lwf,rebuffi2017icarl}. Rigidity refers to the ability of the model not to diverge and remember old knowledge, while plasticity represents the acquisition of new knowledge by DNNs~\cite{de2021continual}. Some pioneering works have attempted to tackle the CL problem in the medical domain~\cite{derakhshani2022lifelonger}. Catastrophic forgetting may occur in various real-world medical scenarios, e.g., data collected (i) over time, and (ii) across devices/institutions. More seriously, due to issues of data privacy, storage, and licensing, old data may not be accessible anymore~\cite{lee2020clinical}. Therefore, it is necessary to develop a non-exemplar CL method for surgical VQLA tasks to resist catastrophic forgetting in clinical applications. 

\begin{figure*}[t]
    \centering
    \includegraphics[width=1\linewidth, trim=-10 560 560 0]{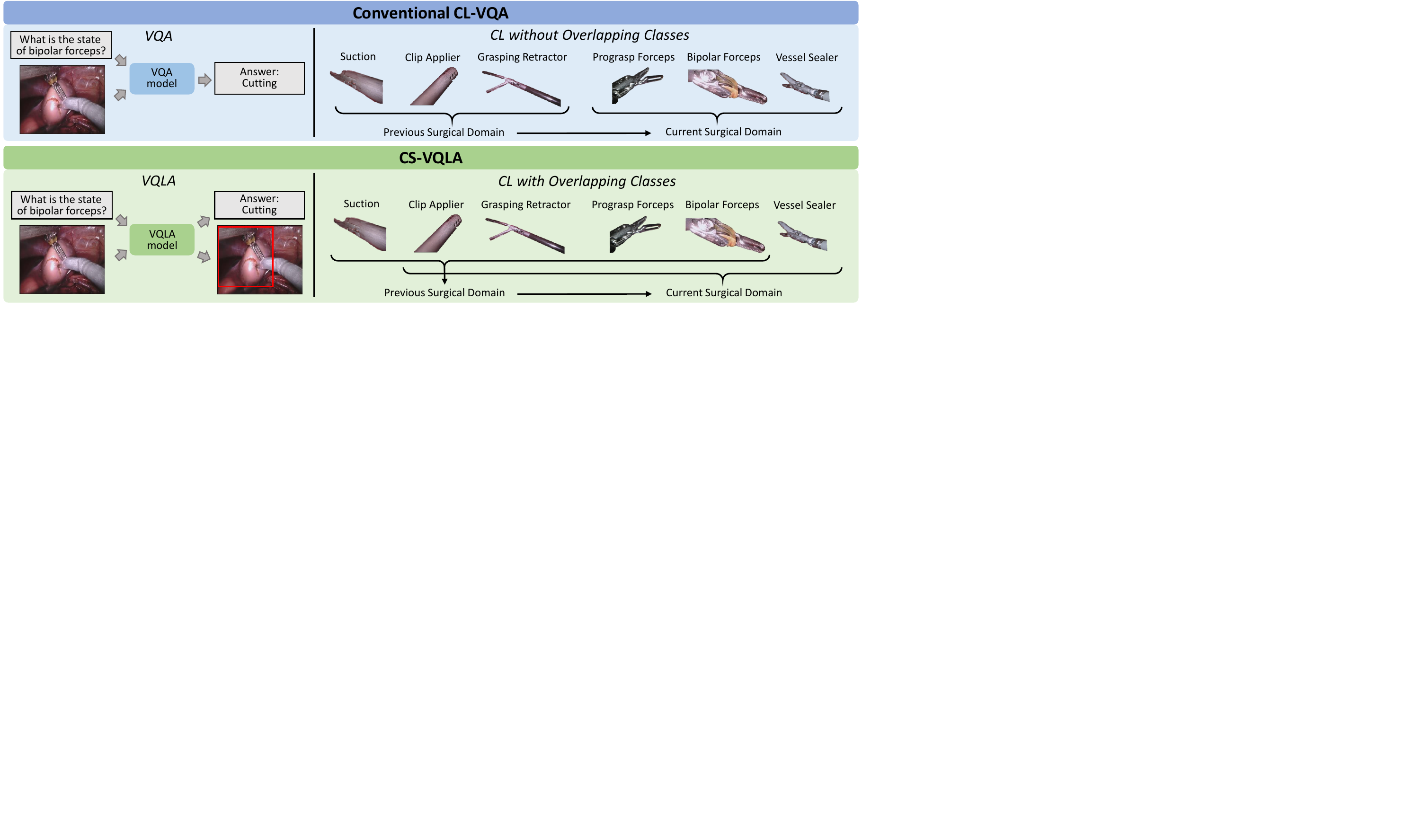}
    \caption{
    Comparison between conventional CL-VQA and our CS-VQLA. Besides providing localized answers, our CS-VQLA framework also pays attention to the issue of overlapping and non-overlapping classes in sequential surgical domains.
    }
    \label{fig:visualization}
\end{figure*}

Furthermore, most medical decision-making tasks shall include classes overlapping with the old tasks and newly appeared classes, as shown in Fig.~\ref{fig:visualization}. We should not distillate the entire previous model when we deal with CL with overlapping classes. Firstly, the model will not emphasize new classes and have a high bias toward overlapping classes rather than new classes. Overlapping classes will dominate the model prediction if we naively follow the distillation from existing CL models. Secondly, catastrophic forgetting will be severe in old non-overlapping classes and the overlapping classes will dominate in the model prediction, and forget the old classes. For this purpose, we revisit distillation methods in CL and design a Continual Surgical VQLA (CS-VQLA) framework for learning incremental classes by balancing the performance of the old overlapping and non-overlapping classes. CS-VQLA has the following attributes: (i) it is a multi-task model including answering and localization, (ii) domain shift and class increment problems both exist, (iii) there may be overlapping classes between old and new tasks. These points shall further complicate the CL tasks.

In this work, \textbf{(1)} We establish a non-exemplar CS-VQLA framework. While being applied to surgical education and scene understanding, the framework can learn data in a streaming manner and effectively resist catastrophic forgetting. \textbf{(2)} We revisit the distillation method for CL, and propose rigidity-plasticity-aware distillation (RP-Dist) and self-calibrated heterogeneous distillation (SH-Dist) for the output logits and intermediate feature maps, respectively. The weight aligning (WA) technique is further integrated to adjust model bias between old and new data. \textbf{(3)} Extensive comparison and ablation studies prove the outstanding performance of our method in mitigating catastrophic forgetting, demonstrating its potential in real-world applications.

\section{Methodology}

\subsection{Preliminaries}
\subsubsection{Problem Definition}
We define the continual learning sequence with $\mathcal{TP}$ time periods, and $t \in \{1,...,\mathcal{TP}\}$ means the current time period. $\mathcal{D}_t$ denotes the training dataset at time period $t$, with $x$ representing a sample of the input question and image pair in $\mathcal{D}_t$. $\mathcal{C}_{old}$ denotes the classes appearing in previous time period $\{1,...,t-1\}$, and $\mathcal{C}_{new}$ represents the classes appearing in current time period $t$. Furthermore, we define the classes existing in both $\mathcal{C}_{old}$ and $\mathcal{C}_{new}$ as \textit{overlapping classes} $\mathcal{C}_{op}$, and define unique classes in $\mathcal{C}_{old}$ as \textit{old non-overlapping classes} $\mathcal{C}_{no}$. $F$ stands for the output feature map from the network backbone. 

\subsubsection{Knowledge Distillation} (KD)~\cite{hinton2015distilling, yuan2020revisiting} on output logits~\cite{li2017lwf} or intermediate feature map~\cite{michieli2019incremental} is a widely used approach to retain knowledge on old tasks. With $z^o$ and $z^{cl}$ denote the output logits from the old and CL model, respectively, we can formulate the logits distillation loss~\cite{li2017lwf} as:
\begin{equation}
\mathcal{L}_{LKD}=\sum_{c=0}^{\mathcal{C}_{old}}-p^o_T(x) \log \left(p^{cl}_T(x)\right)
\label{equ:lwf_loss}
\end{equation}
in which $p^o_T(x) = SM(z^o/T)$ and $p^{cl}_T(x)= SM(z^{cl}/T)$ represent the probabilities. $SM$ means Softmax. $T$ is temperature normalization for all old classes.

\subsubsection{Weight Aligning} (WA)~\cite{zhao2020wa} is a simple technique to align the weight bias in the classifier layer. We use $\mathbf{W}_{new}$ to represent the weights for newly appeared classes in the classifier, and $\mathbf{W}_{old}$ to denote those of old classes, then we have: 
\begin{equation}
\hat{\mathbf{W}}_{new}=\frac{Mean[Norm(\mathbf{W}_{old})]}{Mean[Norm(\mathbf{W}_{new})]} \cdot \mathbf{W}_{new}
\end{equation}
where $norm$ means normalizing all the elements in the vector. In class-incremental learning, WA can effectively avoid the model bias towards new classes.

\subsection{Continual Surgical VQLA (CS-VQLA)}

\begin{figure*}[t]
    \centering
    \includegraphics[width=\linewidth, trim=0 160 430 0]{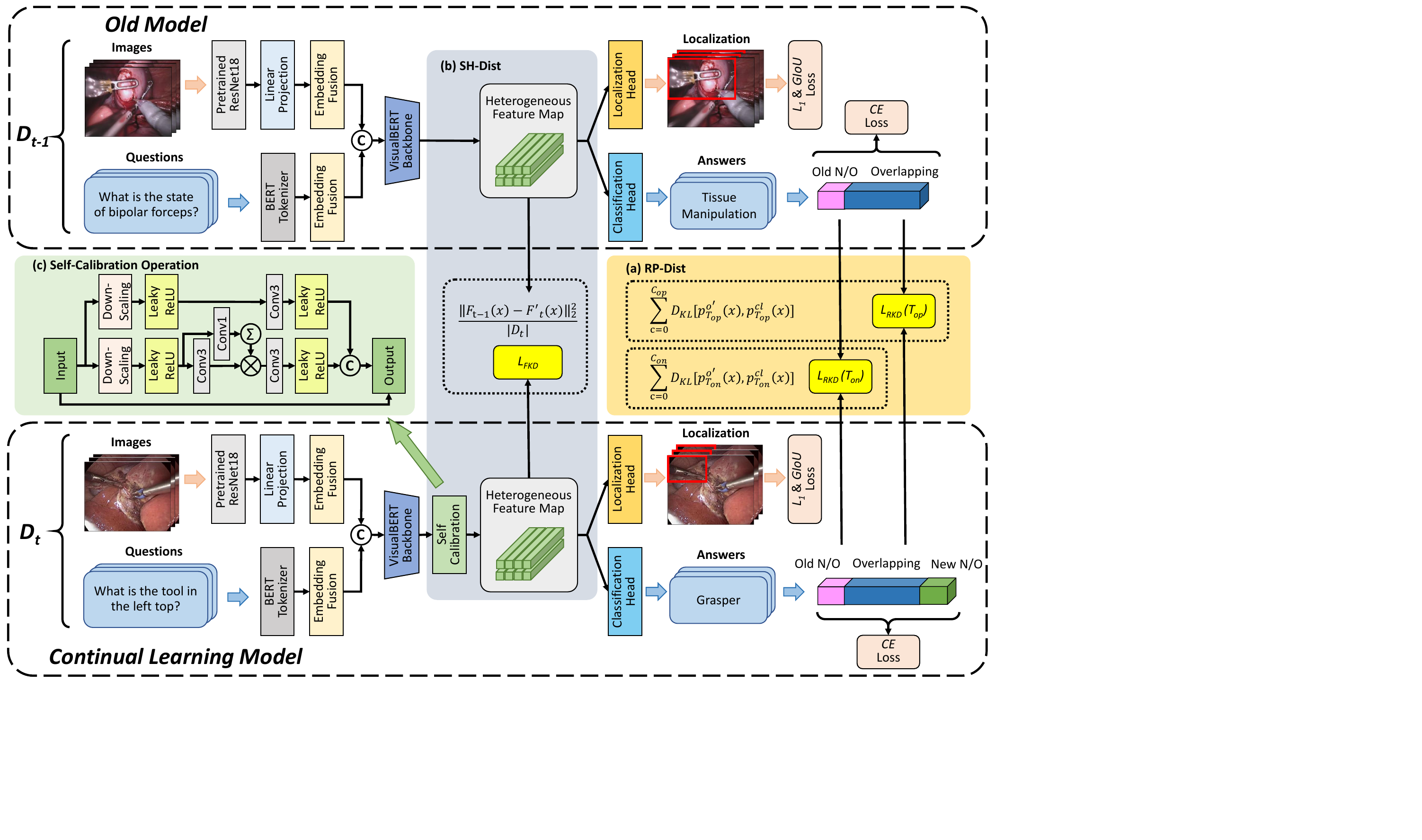}
    \caption{
    Overview of our CS-VQLA network. The VQLA model is used to process bimodal input (image and text) and provide predictions for two tasks (answering and localization). The proposed RP-Dist and SH-Dist are designed to help the CL model retain old knowledge from the old model and trade-off model rigidity-plasticity. `N/O' means non-overlapping classes.}
    \label{fig:main}
\end{figure*}

\subsubsection{Visual-Question Localized Answering} We define our VQLA framework following~\cite{bai2023surgical}, by building a parallel detector on top of the VQA-based classification model. Therefore, the VLQA model includes the following components: a ResNet18~\cite{he2016resnet} pre-trained on ImageNet~\cite{deng2009imagenet} as a prior image feature extractor, a BERT tokenizer~\cite{devlin2018bert}, the VisualBERT~\cite{li2019visualbert} as the backbone (it can also be called as the deep feature extractor), a fully-connected layer as the classifier, and a 3-layer MLP as the detector. The classification task is optimized via the cross-entropy loss $\mathcal{L}_{CE}$, and the bounding box regression is optimized by the sum of $\mathcal{L}_1$ and $GIoU$ loss~\cite{rezatofighi2019generalized}. Thus, the VQLA loss can be formulated as: $\mathcal{L}_{VQLA}=\mu \cdot \mathcal{L}_{CE}+(\mathcal{L}_1+\mathcal{L}_{GIoU})$,
where $\mu$ is set as $100$ to balance the optimization progress of the two tasks.

\subsubsection{Rigidity-Plasticity-Aware Distillation (RP-Dist)} The current rigidity-plasticity trade-off is towards the entire model. However, we shall make the rigidity-plasticity aware in overlapping and non-overlapping classes. 

There is no overlap between $\mathcal{C}_{old}$ and $\mathcal{C}_{new}$ in an ideal class-incremental learning setup, so the temperature $T$ in Equ.~\ref{equ:lwf_loss} is set to $2$ by~\cite{li2017lwf}.
However, in a real-world application setup, $T$ should not smooth the logits equally for old non-overlapping $\mathcal{C}_{on}$ and overlapping classes $\mathcal{C}_{op}$. 
Specifically, through adjusting for $T$, we shall endow the model greater plasticity on $C_{op}$, and keep the rigidity on $C_{on}$.
We first establish a regularized distillation loss. Originally, the old model shall serve as the `teacher' model in CL-based distillation. Instead of directly distilling the old model output logits, we construct a perfect pseudo teacher for distillation. To begin with, a pseudo answering label set $a'$ can be built from the old model classification probabilities $p^{o}(x)$ via $a' = Max[p(x)]$. Based on the idea of label smoothing, we can manually setup a pseudo old model to have a high probability of predicting a correct class, and its probability distribution shall be:
\begin{equation} 
    {p^o}'(x)= \begin{cases}\lambda & \text { if } x=a' \\ (1-\lambda) /(\mathcal{C}_{old}-1) & \text { if } x \neq a'\end{cases}
\end{equation}
When $\lambda$ is set to a very high number (e.g., $\lambda \geq 0.9$), we will have a high probability of getting a correct class, allowing the teacher model to have perfect performance. The probability output of the CL model can be optimized with this pseudo-teacher based on the Kullback-Leibler divergence $D_{KL}$:
\begin{equation}
    \mathcal{L}_{RKD}=\sum_{c=0}^{\mathcal{C}_{old}}D_{K L}\left[p_T^{o'}(x), p^{cl}_T(x)\right]
\end{equation}
$T$ is the KD temperature used to generate soft probabilities for the pseudo old model. As discussed above, this naive setting of $T$ is not suitable for general CL scenarios. Therefore, we treat $T_{op}$ and $T_{on}$ differently to strengthen the plasticity on $\mathcal{C}_{op}$ and the rigidity on $\mathcal{C}_{on}$ respectively. $\mathcal{L}_{RKD}$ can thereby be rewritten as:
\begin{equation}
\begin{aligned}  
\mathcal{L}_{RKD}=\sum_{c=0}^{\mathcal{C}_{op}}D_{K L}\left[p_{T_{op}}^{o'}(x), \, p^{cl}_{T_{op}}(x)\right] + \sum_{c=0}^{\mathcal{C}_{on}}D_{K L}\left[p_{T_{on}}^{o'}(x), \, p^{cl}_{T_{on}}(x)\right]
\end{aligned}
\end{equation}
We keep $T_{op} > T_{on}$ to balance the rigidity and plasticity trade-off in the CL model, and set $T_{op} = 25$, $T_{on} = 20$ empirically in our implementation. 

\subsubsection{Self-Calibrated Heterogeneous Distillation (SH-Dist)} Works have 
discussed the use of self-calibration to improve model performance~\cite{liu2020improving, zou2022self}. However, assuming we obtain an old model and we would like to conduct CL training on it, we can hardly modify the old model itself directly. Therefore, we perform a self-calibration operation on the heterogeneous output features $F_t$ from the VisualBERT backbone to get self-calibrated feature $F'_t$. The details can be referred to at the bottom of Fig.~\ref{fig:main}. Without engaging more learnable parameters, we endow the heterogeneous features with adaptively modeled long-range context information. 
Therefore, we can construct our feature distillation using the self-calibrated feature map $F'_t$  and the old model feature map $F_{t-1}$. $\mathcal{L}_2$ loss is used to minimize the distance between $F'_t$ and $F_{t-1}$ empirically by following~\cite{michieli2019incremental}:
\begin{equation}
    \mathcal{L}_{FKD} = \frac{{\|F_{t-1}(x) - F'_t(x)\|}^2_2}{\left|\mathcal{D}_t\right|}
\end{equation}

Subsequently, the self-calibrated feature map $F'_t$ shall be propagated through the parallel classifier and detector for the multi-task prediction. 

\subsubsection{Overall Framework} Fig.~\ref{fig:main} shows the overview of our CS-VQLA framework. The given image and question input are respectively processed as feature embedding by pre-trained ResNet18 and BERT tokenier, and fed to the VisualBERT backbone after embedding fusion. Then the output heterogeneous feature map is used to train the parallel predictors. The loss functions establish the essential components of our CS-VQLA framework. In the initial time period $t=0$, the model is only trained on the VQLA loss. When $t > 0$, we combine the VQLA loss for training on the current dataset $D_t$, with the RP-Dist \& SH-Dist loss to retain the old knowledge. We can summarize our final loss function as follows:
\begin{equation}
\mathcal{L}=\left\{
\begin{aligned}
& \mathcal{L}_{VQLA} & t=0 \\
& \alpha \cdot \mathcal{L}_{VQLA} + \beta \cdot \mathcal{L}_{RKD} + \gamma \cdot \mathcal{L}_{FKD} \; & t > 0
\end{aligned}
\right.
\end{equation}
We set $\alpha = \beta = 1$ and $\gamma = 5$ in our implementation. Furthermore, WA is deployed after training on each time period, to balance the weight bias of new classes on the classification layer. Through the combination of multiple distillation paradigms and model weight adjustment, we successfully realize the general continual learning framework in the VQLA scenario.

\section{Experiments}
\subsection{Dataset and Setup}
\subsubsection{Dataset} We construct our continual procedure as follows: when $t=0$, we train on EndoVis18 Dataset, $t=1$ on EndoVis17 Dataset, and $t=2$ on M2CAI Dataset. Therefore, we can establish our CS-VQLA framework with a large initial step, and several smaller sequential steps. When splitting the dataset, we isolate the training and test sets in different sequences to avoid information leakage.

\textit{EndoVis18 Dataset} is a public dataset with 14 videos on robotic surgery~\cite{allan2020endovis18}. The question-answer (QA) pairs are accessible in~\cite{seenivasan2022surgical}, and the bounding box annotations are from~\cite{islam2020learning}. The answers are in single-word form with three categories (organ, interaction, and locations). We further extend the QA pairs and include cases when the answer is a surgical tool. Besides, if the answer is regarding the organ-tool interaction, the bounding box shall contain both the organ and the tool. Statistically, the training set contains 1560 frames with 12741 QA pairs, and the test set contains 447 frames with 3930 QA pairs. 

\textit{EndoVis17 Dataset} is a public dataset with 10 videos on robotic surgery~\cite{allan2019endovis17}. We randomly select frames and manually annotate the QA pairs and bounding boxes. The training set contains 167 frames with 1034 QA pairs, and the test set contains 40 frames with 201 QA pairs. 

\textit{M2CAI Dataset} is also a public robotic surgery dataset~\cite{stauder2016tum, twinanda2016endonet}, and the location bounding box is publicly accessible in~\cite{jin2018tool}. Similarly, we randomly select 167 frames and annotate 449 QA pairs for the training set, and 40 frames with 94 QA pairs in different videos for the test set.
\\
\\
\textbf{Implementation Details}
We compare our solution against the fine-tuning (FT) baseline and state-of-the-art (SOTA) methods, including LwF~\cite{li2017lwf}, WA~\cite{zhao2020wa}, iCaRL~\cite{rebuffi2017icarl}, IL2A~\cite{zhu2021il2a}, PASS~\cite{zhu2021pass}, SSRE~\cite{zhu2022ssre}, CLVQA~\cite{lei2023symbolic}, and CLiMB~\cite{srinivasan2022climb}. All the methods are implemented using~\cite{zhou2021pycil}\footnote{\href{https://github.com/G-U-N/PyCIL}{github.com/G-U-N/PyCIL}}, with PyTorch and on NVIDIA RTX 3090 GPU. We removed the exemplars in all methods for a non-exemplar comparison. All methods are firstly trained on EndoVis18 ($t=0$) for $60$ epochs with a learning rate of $1 \times 10^{-5}$, and then trained on EndoVis17 ($t=2$) and M2CAI ($t=2$) for $30$ epochs with a learning rate of $5 \times 10^{-5}$. We use Adam optimizer~\cite{kingma2014adam} and a batch size of $64$. Answering and localization performance are evaluated by Accuracy (Acc) and mean intersection over union (mIoU), respectively.

\begin{table}[t]
    \caption{
    Comparison experiments from the time period $t=0$ to $t=1$. \textbf{Bold} and \underline{underlined} represent best and second best, respectively. `W/O' denotes `without', and `W/I' denotes `within'. `N/O' means non-overlapping. `Old N/O' represents the classes that exist in $t=0$ but do not exist in $t=1$, and `New N/O' represents the opposite. `Overlapping' denotes the classes that exist in both $t=0,1$.}
 	\centering
	\label{tab:main0-1}  
 \resizebox{\textwidth}{!}{	
\begin{tabular}{c|c|cc|cc|cc|cc|cc|cc}
\hline
\multirow{2}{*}{$t=1$} &\multirow{2}{*}{Methods} & \multicolumn{2}{c|}{Old N/O} & \multicolumn{2}{c|}{Overlapping} & \multicolumn{2}{c|}{New N/O} &  \multicolumn{2}{c|}{EndoVis18} & \multicolumn{2}{c|}{EndoVis17} & \multicolumn{2}{c}{Average} \\ \cline{3-14} 
& & Acc & mIoU & Acc & mIoU & Acc & mIoU & Acc & mIoU & Acc & mIoU & Acc & mIoU  \\ \hline
\multirow{2}{*}{W/O CL} & Base ($t=0$) & 6.11 & 62.12 & 64.60 & 75.68 & \XSolidBrush & \XSolidBrush & 62.65 & 75.23 & \XSolidBrush & \XSolidBrush & \XSolidBrush & \XSolidBrush\\
& FT & 0.00 & 62.55 & 38.07 & 71.88 & 86.96 & 80.41 & 34.86 & 71.29 & 81.59 & 78.30 & 58.23 & 74.79 \\\hline
\multirow{9}{*}{W/I CL} & LwF~\cite{li2017lwf} & 0.00 & \textbf{63.40} & 54.36 & 69.94 & \underline{73.91} & 77.47 & 53.20 & 69.53 & 43.79 & 74.64 & 48.50 & 72.08 \\
& WA~\cite{zhao2020wa} & \underline{0.76} & 60.70 & 55.11 & 71.61 & 52.17 & 78.10 & 52.85 & 71.02 & 63.68 & 76.71 & 58.26 & 73.86\\
& iCaRL~\cite{rebuffi2017icarl} & \underline{0.76} & \underline{62.23} & \underline{55.87} & 72.36 & 43.48 & \textbf{79.51} & \underline{53.85} & 71.76 & 58.21 & \textbf{78.41} & 56.03 & \underline{75.08}\\
& IL2A~\cite{zhu2021il2a} & 0.00 & 57.74 & 53.00 & 69.88 & 56.52 & 78.20 & 51.48 & 69.23 & 48.76 & 75.75 & 50.12 & 72.49 \\
& PASS~\cite{zhu2021pass} & 0.00 & 56.49 & 54.01 & 70.08 & 69.57 & 77.60 & 51.70 & 69.46 & 65.67 & 74.56 & 58.69 & 72.01\\
& SSRE~\cite{zhu2022ssre} & 0.00 & 60.29 & 54.04 & 70.34 & 65.22 & 76.07 & 51.76 & 69.78 & 64.68 & 75.44 & 58.22 & 72.61 \\
& CLVQA~\cite{lei2023symbolic} & 0.00 & 59.87 & 51.83 & 72.98 & 65.22 & 78.36 & 49.14 & 72.40 & 72.14 & 76.40 & 60.64 & 74.40\\
& CLiMB~\cite{srinivasan2022climb} & 0.00 & 60.16 & 52.88 & \underline{72.99} & 69.57 & 77.37 & 50.13 & \underline{72.44} & \underline{74.13} & 75.87 & \underline{62.13} & 74.16 \\
& Ours & \textbf{1.53} & 61.08 & \textbf{56.98} & \textbf{74.57} & \textbf{78.26} & \underline{78.59} & \textbf{54.33} & \textbf{74.02} & \textbf{75.12} & \underline{77.02} & \textbf{64.73} & \textbf{75.52}  \\ \hline
\end{tabular}}
\end{table}

\begin{table}[t]
	\caption{
        Comparison experiments from the time period $t=0,1$ to $t=2$. `Old N/O' represents the classes that exist in $t=0,1$ but do not exist in $t=2$, and `New N/O' represents the opposite. `Overlapping' denotes the classes that exist in $t=0,1,2$. 
	}
 	\centering
	\label{tab:main1-2}  
 \resizebox{\textwidth}{!}{	
\begin{tabular}{c|c|cc|cc|cc|cc|cc|cc|cc}
\hline
\multirow{2}{*}{$t=2$} &\multirow{2}{*}{Methods} & \multicolumn{2}{c|}{Old N/O} & \multicolumn{2}{c|}{Overlapping} & \multicolumn{2}{c|}{New N/O} &  \multicolumn{2}{c|}{EndoVis18} & \multicolumn{2}{c|}{EndoVis17} & \multicolumn{2}{c|}{M2CAI16}  & \multicolumn{2}{c}{Average} \\ \cline{3-16} 
& & Acc & mIoU & Acc & mIoU & Acc & mIoU & Acc & mIoU & Acc & mIoU & Acc & mIoU & Acc & mIoU  \\ \hline
W/O CL & FT & 4.00 & 60.90 & 19.08 & 58.69 & 55.56 & 70.83 & 15.57 & 58.75 & 41.79 & 60.18 & 51.06 & 69.48 & 36.14 & 62.80 \\\hline
\multirow{9}{*}{W/I CL} & LwF~\cite{li2017lwf} & 4.20 & \underline{62.91} & \textbf{42.75} & 63.34 & 27.78 & 72.68 & \underline{38.04} & 63.25 & 41.29 & \textbf{62.71} & 31.91 & 69.65 & 37.08 & \underline{65.20} \\
& WA~\cite{zhao2020wa} & 6.80 & 59.32 & 40.62 & 61.55 & \textbf{55.56} & 73.06 & 36.67 & 61.20 & 36.32 & 60.86 & 40.43 & 69.80 & 37.81 & 63.96\\
& iCaRL~\cite{rebuffi2017icarl} & 2.00 & 58.55 & 38.06 & 58.59 & 41.67 & 73.42 & 33.46 & 58.30 & 38.81 & 61.12 & 38.30 & \textbf{71.01} & 36.85 & 63.48\\
& IL2A~\cite{zhu2021il2a} & 11.00 & 57.25 & 33.80 & 58.50 & 27.78 & 72.71 & 30.61 & 58.28 & 37.31 & 58.29 & 36.17 & 67.10 & 34.70 & 61.22 \\
& PASS~\cite{zhu2021pass} & 22.60 & 56.57 & 24.80 & 58.39 & 30.56 & 72.07 & 23.52 & 58.07 & 37.81 & 58.91 & \textbf{41.49} & 66.68 & 34.27 & 61.22\\
& SSRE~\cite{zhu2022ssre} & 13.80 & 58.12 & 19.49 & 57.02 & \underline{47.22} & 73.31 & 18.12 & 57.00 & 27.36 & 58.09 & 40.43 & 67.47 & 28.64 & 60.85 \\
& CLVQA~\cite{lei2023symbolic} & 21.80 & 58.01 & 36.54 & 62.92 & 25.00 & \underline{74.09} & 34.40 & 62.35 & 39.80 & 61.05 & 36.17 & 68.89 & 36.79 & 64.10 \\
& CLiMB~\cite{srinivasan2022climb} & \underline{23.00} & 57.03 & 38.90 & \underline{64.30} & 33.33 & 73.45 & 36.62 & \underline{63.50} & \underline{42.29} & 61.35 & 40.43 & 69.20 & \underline{39.78} & 64.68 \\
& Ours & \textbf{28.20} & \textbf{68.14} & \underline{41.04} & \textbf{65.74} & 44.44 & \textbf{74.41} & \textbf{39.13} & \textbf{66.21} & \textbf{46.77} & \underline{62.17} & \textbf{41.49} & \underline{70.04} & \textbf{42.46} & \textbf{66.14} \\ \hline
\end{tabular}}
\end{table}

\subsection{Results}

Except for testing on three datasets separately, we set three specific categories in our continual learning setup: \textit{old non-overlapping} (old N/O) classes, \textit{overlapping} classes, and \textit{new non-overlapping} (new N/O) classes. By measuring the performance in these three categories, we can easily observe the catastrophic forgetting phenomenon and the performance of mitigating catastrophic forgetting. 

As shown in Table~\ref{tab:main0-1}~\&~\ref{tab:main1-2}, firstly, catastrophic forgetting can be apparently observed in the performance of FT. Then, among all baselines, iCaRL achieves the best performance when the model learns from $t=0$ to $t=1$, and gets to forget when there are more time periods. On the contrary, LwF exhibits a strong retention of old knowledge, but a lack of ability to learn new. Our proposed methods demonstrate superior performance in almost all metrics and classes. In classification tasks, the overall average of our methods outperforms the second best with 2.60\% accuracy improvement at $t=1$ and 2.68\% at $t=2$. In localization tasks, our method is 0.44 mIoU higher than the second best at $t=1$ and 0.94 mIoU higher at $t=2$. The results prove the remarkable ability of our method to balance the rigidity-plasticity trade-off. Furthermore, an ablation study is conducted to demonstrate the effectiveness of each component in our proposed method. We (i) degenerate the RP-Dist to original logits distillation~\cite{li2017lwf}, (ii)~degenerate the SH-Dist to normal feature distillation~\cite{michieli2019incremental}, and (iii) remove the WA module, as shown in Table~\ref{tab:abl-1}. Experimental results show that each component we propose or integrate plays an essential role in the final rigidity-plasticity trade-off. Therefore, we demonstrate that each of our components is indispensable. More evaluation and ablation studies can be found in the supplementary materials.
\begin{table}[t]
	\caption{
        Ablation experiments from the time period $t=0$ to $t=1$, and from $t=1$ to $t=2$. To observe the contribution of each component, we degenerate the proposed RP-Dist and SH-Dist to the normal distillation paradigm, and remove the WA module.
	}
 	\centering
	\label{tab:abl-1}  
 \resizebox{\textwidth}{!}{	
\begin{tabular}{ccc|cc|cc|cc|cc|cc|cc|cc|cc}
\hline
 \multicolumn{3}{c|}{Methods} & \multicolumn{8}{c|}{$t=0$ to $t=1$} & \multicolumn{8}{c}{$t=1$ to $t=2$}\\ \cline{4-19}
 \multicolumn{3}{c|}{} & \multicolumn{2}{c|}{Old N/O} & \multicolumn{2}{c|}{Overlapping} & \multicolumn{2}{c|}{New N/O} & \multicolumn{2}{c|}{Average} & \multicolumn{2}{c|}{Old N/O} & \multicolumn{2}{c|}{Overlapping} & \multicolumn{2}{c|}{New N/O} & \multicolumn{2}{c}{Average}\\ \hline 
 RP & SH & WA & Acc & mIoU & Acc & mIoU & Acc & mIoU & Acc & mIoU & Acc & mIoU & Acc & mIoU & Acc & mIoU & Acc & mIoU  \\ \hline
\Checkmark & \XSolidBrush & \XSolidBrush & 0.00 & 60.58 & 55.30 & 72.94 & 73.91 & 77.62 & 62.66 & 74.37 & 11.40 & 58.29 & 40.96 & 64.72 & 30.56 & 73.31 & 40.80 & 64.85 \\
\XSolidBrush & \Checkmark & \XSolidBrush & 0.00 & 59.94 & 56.23 & 73.66 & \textbf{82.61} & 78.04 & 64.33 & 74.90 & 9.80 & 57.17 & 38.14 & 65.30 & 38.89 & 74.15 & 40.26 & 64.35 \\
\XSolidBrush & \XSolidBrush & \Checkmark & 0.00 & 60.33 & 53.08 & 72.45 & 52.17 & 76.18 & 61.94 & 74.30 & 12.20 & 59.49 & 39.75 & 64.81 & 41.67 & 72.12 & 39.07 & 64.81 \\
\XSolidBrush & \Checkmark & \Checkmark & 0.00 & 60.97 & 54.59 & 74.12 & 73.91 & 78.52 & 61.83 & 74.89 & 11.00 & 65.16 & 40.69 & 64.45 & 41.67 & 74.18 & 39.63 & 65.45 \\
\Checkmark & \XSolidBrush & \Checkmark & 0.00 & 61.04 & 56.08 & 74.11 & 60.87 & 78.91 & 61.60 & 74.88 & 11.00 & 59.10 & 39.25 & 62.60 & 22.22 & 71.66 & 39.16 & 64.55 \\
\checkmark & \checkmark & \XSolidBrush & 0.00 & 60.07 & 54.56 & 74.15 & 73.91 & \textbf{79.79} & 61.81 & 75.00 & 10.40 & 63.32 & 39.14 & 64.40 & 27.78 & 73.89 & 40.21 & 65.65 \\
\Checkmark & \Checkmark & \Checkmark & \textbf{1.53} & \textbf{61.08} & \textbf{56.98} & \textbf{74.57} & 78.26 & 78.59 & \textbf{64.73} & \textbf{75.52} & \textbf{28.20} & \textbf{68.14} & 41.04 & \textbf{65.74} & \textbf{44.44} & \textbf{74.41} & \textbf{42.46} & \textbf{66.14}\\ \hline
\end{tabular}}
\end{table}

\section{Conclusion}
This paper introduces CS-VQLA, a general continual learning framework on surgical VQLA tasks. This is a significant attempt to continue learning under complicated clinical tasks. Specifically, we propose the RP-Dist on output logits, and the SH-Dist on the intermediate feature space, respectively. The WA technique is further integrated for model weight bias adjustment. Superior performance on VQLA tasks demonstrates that our method has an excellent ability to deal with CL-based surgical scenarios. Except for giving localized answers for better surgical scene understanding, our solution can conduct continual learning in any questions in surgical applications to solve the problem of class increment, domain shift, and overlapping/non-overlapping classes. Our framework can also be applied when adapting a vision-language foundation model in the surgical domain. Therefore, our solution holds promise for deploying auxiliary surgical education tools across time/institutions. Potential future works also include combining various surgical training systems (e.g., mixed reality-based training, surgical skill assessment) to develop an effective and comprehensive virtual teaching system.

\subsubsection{Acknowledgements.}

This work was funded by Hong Kong RGC CRF C4063-18G,  CRF C4026-21GF, RIF R4020-22, GRF 14203323, GRF 14216022, GRF 14211420, NSFC/RGC JRS N\_CUHK420/22; Shenzhen-Hong Kong-Macau Technology Research Programme (Type C 202108233000303); Guangdong GBABF \#2021B1515120035. M. Islam was funded by EPSRC grant [EP/W00805X/1].

\bibliography{reference}
\bibliographystyle{splncs04}

\newpage

\section*{Supplementary Materials for ``Revisiting Distillation for Continual Learning on Visual Question Localized-Answering in Robotic Surgery''}

\begin{figure}
\centering
\includegraphics[width=\textwidth, trim=0 620 660 0]{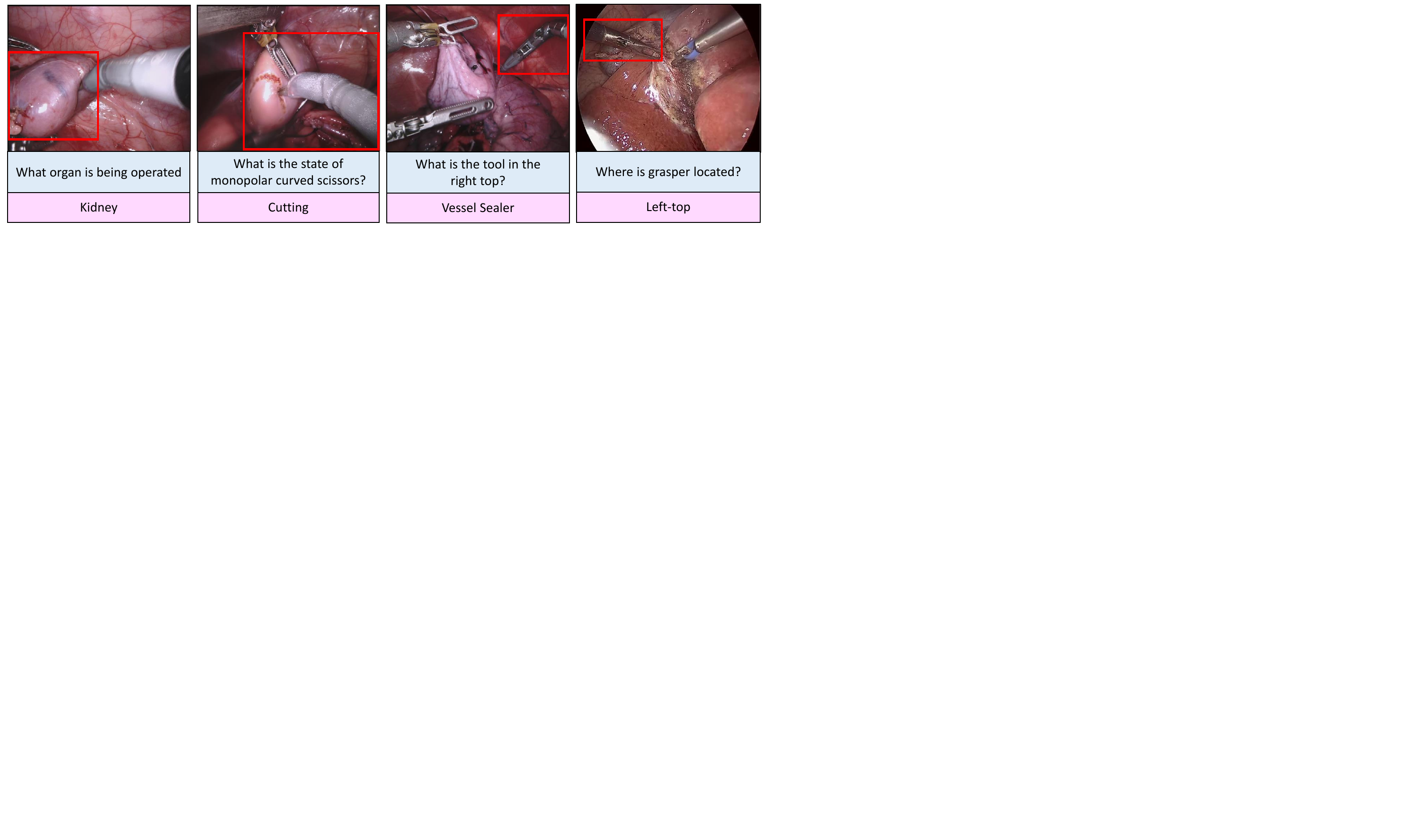}

\caption{Visualization of the datasets used in this study.} \label{figa1}
\end{figure}

\begin{table}[!h]
	\caption{
Overlapping and non-overlapping class distributions in EndoVis18, EndoVis17 \& M2CAI datasets.
	}
 	\centering
	\label{tab:abl-1}  
 \resizebox{\textwidth}{!}{	
\begin{tabular}{c|cccccccccccccc}
\hline
Time Periods & kidney & Idle & Grasping & Retraction & \makecell[c]{Tissue \\ Manipulation} & \makecell[c]{Tool \\ Manipulation} & Cutting & Cauterization & Suction & Looping & Suturing  \\ \hline
EndoVis18 & \Checkmark & \Checkmark & \Checkmark & \Checkmark & \Checkmark & \Checkmark & \Checkmark & \Checkmark & \Checkmark & \Checkmark & \Checkmark \\
EndoVis17 & \XSolidBrush & \Checkmark & \Checkmark & \Checkmark & \Checkmark & \Checkmark & \Checkmark & \Checkmark & \Checkmark & \Checkmark & \Checkmark \\
M2CAI & \XSolidBrush & \Checkmark & \Checkmark & \XSolidBrush & \Checkmark & \XSolidBrush & \Checkmark & \Checkmark & \XSolidBrush & \XSolidBrush & \Checkmark \\ \hline
Time Periods & Clipping & \makecell[c]{Ultrasound \\ Sensing} & Staple & \makecell[c]{bipolar} & scissors & \makecell[c]{prograsp \\ forceps} & \makecell[c]{large needle \\ driver} & \makecell[c]{ultrasound \\ probe} & \makecell[c]{clip \\ applier} &  suction & stapler \\ \hline
EndoVis18 & \Checkmark & \Checkmark & \Checkmark & \Checkmark & \Checkmark & \Checkmark & \Checkmark & \Checkmark & \Checkmark & \Checkmark & \Checkmark \\
EndoVis17 & \Checkmark & \Checkmark & \Checkmark & \Checkmark & \Checkmark & \Checkmark & \Checkmark & \Checkmark & \XSolidBrush & \XSolidBrush & \XSolidBrush \\
M2CAI & \Checkmark & \XSolidBrush & \XSolidBrush & \Checkmark & \Checkmark & \XSolidBrush & \XSolidBrush & \XSolidBrush & \XSolidBrush & \XSolidBrush & \XSolidBrush \\ \hline
Time Periods  & \makecell[c]{grasping \\ retractor} & \makecell[c]{vessel\\sealer} & irrigator & hook & grasper & specimenbag & clipper & \makecell[c]{left-top} & \makecell[c]{right-top} & \makecell[c]{left-\\bottom} & \makecell[c]{right-\\bottom}\\ \hline
EndoVis18 & \XSolidBrush & \XSolidBrush & \XSolidBrush & \XSolidBrush & \XSolidBrush & \XSolidBrush & \XSolidBrush & \Checkmark & \Checkmark & \Checkmark & \Checkmark \\
EndoVis17 & \Checkmark & \Checkmark & \XSolidBrush & \XSolidBrush & \XSolidBrush & \XSolidBrush & \XSolidBrush & \Checkmark & \Checkmark & \Checkmark & \Checkmark \\
M2CAI & \XSolidBrush & \XSolidBrush & \Checkmark & \Checkmark & \Checkmark & \Checkmark & \Checkmark & \Checkmark & \Checkmark & \Checkmark & \Checkmark \\ \hline
\end{tabular}}
\end{table}

\begin{table}[!h]
	\caption{
	Ablation experiments on temperature normalization setup of RP-Dist from the time period $t=0$ to $t=1$, and from $t=1$ to $t=2$. We finetune the $T_{op}$ \& $T_{on}$ to compare the performance.
	}
 	\centering
	\label{tab:abl-1}  
 \resizebox{\textwidth}{!}{	
\begin{tabular}{cc|cc|cc|cc|cc|cc|cc|cc|cc}
\hline
 \multicolumn{2}{c|}{Time Periods}& \multicolumn{8}{c|}{$t=0$ to $t=1$} & \multicolumn{8}{c}{$t=1$ to $t=2$}\\ \hline 
 \multicolumn{2}{c|}{Temperature} & \multicolumn{2}{c|}{Old N/O} & \multicolumn{2}{c|}{Overlapping} & \multicolumn{2}{c|}{New N/O} & \multicolumn{2}{c|}{Average} & \multicolumn{2}{c|}{Old N/O} & \multicolumn{2}{c|}{Overlapping} & \multicolumn{2}{c|}{New N/O} & \multicolumn{2}{c}{Average}\\ \hline 
 $T_{op}$ &  $T_{no}$ & Acc & mIoU & Acc & mIoU & Acc & mIoU & Acc & mIoU & Acc & mIoU & Acc & mIoU & Acc & mIoU & Acc & mIoU  \\ \hline
15 & 10 & 0.00 & 59.06 & 54.36 & 74.38 & 34.78 & 76.82 & 62.31 & 74.80 & 21.60 & 66.23 & 40.37 & 63.80 & 38.89 & 72.47 & 38.24 & 65.15\\
20 & 10 & 0.76 & 60.60 & 55.29 & 74.04 & 60.87 & \underline{78.11} & 63.57 & 75.04 & 23.40 & 66.63 & 40.88 & \underline{65.35} & 30.56 & 73.50 & 41.02 & 65.68\\
20 & 15 & 0.76 & \textbf{61.70} & \underline{56.29} & 74.12 & 52.17 & 74.67 & 62.40 & 75.18 & 24.00 & 66.21 & 40.58 & 64.11 & 30.56 & 73.13 & 41.59 & 64.71\\
25 & 10 & \textbf{1.53} & 60.95 & 55.34 & 73.77 & 60.87 & 78.05 & 63.61 & 74.87 & 26.00 & 67.31 & 36.79 & 65.27 & \textbf{47.22} & 73.19 & 41.02 & 65.68 \\
25 & 15 & 0.76 & 60.02 & 56.10 & 74.66 & \underline{78.26} & 77.38 & \underline{64.03} & \underline{75.39} & 23.80 & 65.74 & \textbf{41.22} & 64.92 & 30.56 & 73.26 & \underline{42.38} & 65.60\\
25 & 20 & \textbf{1.53} & \underline{61.08} & \textbf{56.98} & \underline{74.57} & \underline{78.26} & \textbf{78.59} & \textbf{64.73} & \textbf{75.52} & \textbf{28.20} & \textbf{68.14} & 41.04 & \textbf{65.74} & \underline{44.44} & \textbf{74.41} & \textbf{42.46} & \textbf{66.14}\\
30 & 10 & \textbf{1.53} & 60.76 & 56.06 & 74.15 & 60.87 & 76.99 & 62.56 & 75.04 & \underline{27.60} & 67.37 & 39.04 & 65.17 & 27.78 & 73.45 & 40.91 & 65.93 \\
30 & 15 & 0.76 & 59.63 & 54.71 & 74.39 & 73.91 & 76.82 & 62.61 & 74.95 & 20.40 & \underline{67.76} & 38.41 & 64.59 & 36.11 & 71.12 & 41.95 & 65.51\\
30 & 20 & 0.76 & 60.92 & 54.77 & 73.78 & \textbf{82.61} & 76.25 & 62.19 & 74.46 & 25.20 & 66.73 & 38.00 & 64.79 & 44.44 & \underline{74.18} & 39.12 & 65.24\\
30 & 25 & 0.00 & 59.15 & 53.46 & \textbf{74.70} & 56.52 & 76.24 & 62.38 & 75.39 & 21.60 & 67.21 & \underline{41.07} & 65.09 & 38.89 & 73.04 & 39.70 & \underline{65.98}\\ \hline
\end{tabular}}
\end{table}

\begin{table}[!h]
	\caption{
	Ablation experiments on different ratio of the loss combination from the time period $t=0$ to $t=1$, and from $t=1$ to $t=2$.
	}
 	\centering
	\label{tab:abl-1}  
 \resizebox{\textwidth}{!}{	
\begin{tabular}{ccc|cc|cc|cc|cc|cc|cc}
\hline
 \multicolumn{3}{c|}{Time Periods} & \multicolumn{6}{c|}{$t=0$ to $t=1$} & \multicolumn{6}{c}{$t=1$ to $t=2$}\\ \hline 
 \multicolumn{3}{c|}{Loss Factors} & \multicolumn{2}{c|}{Old N/O} & \multicolumn{2}{c|}{Overlapping} & \multicolumn{2}{c|}{New N/O} & \multicolumn{2}{c|}{Old N/O} & \multicolumn{2}{c|}{Overlapping} & \multicolumn{2}{c|}{New N/O} \\ \hline 
 \multicolumn{1}{c}{\textcolor{white}{1}$\alpha$}\textcolor{white}{1} & \multicolumn{1}{c}{$\beta$} & \multicolumn{1}{c|}{$\gamma$} & Acc & mIoU & Acc & mIoU & Acc & mIoU & Acc & mIoU & Acc & mIoU & Acc & mIoU \\ \hline
1 & 1 & 5 & \textbf{1.53} & \textbf{61.08} & \textbf{56.98} & \textbf{74.57} & \textbf{78.26} & 78.59 & \textbf{28.20} & \textbf{68.14} & 41.04 & \textbf{65.74} & \textbf{44.44} & 74.41 \\
1 & 1 & 10 & \textbf{1.53} & 59.94 & 56.20 & 73.66 & \textbf{78.26} & 76.62 & 26.71 & 62.93 & 37.93 & 65.20 & 38.89 & 73.63 \\
1 & 5 & 5 & 0.00 & 60.98 & 56.05 & 74.40 & 73.91 & \textbf{79.19} & 22.14 & 59.35 & 39.66 & 65.06 & 25.00 & 73.72\\ 
1 & 2.5 & 5 &  0.00 & 60.83 & 56.46 & 74.35 & \textbf{78.26} & 78.22 & 27.48 & 63.53 & \textbf{41.38} & 64.97 & 38.89 & \textbf{74.69}\\ \hline
\end{tabular}}
\end{table}

\begin{table}[t]
	\caption{
		Comparison experiments on the question-answering task from the time period $t=0$ to $t=1$, and from $t=1$ to $t=2$. The performance is  evaluated with Balanced Accuracy. 
	}
 	\centering
	\label{tab:main1-2}  
 \resizebox{\textwidth}{!}{	
\begin{tabular}{c|ccc|ccc}
\hline
\multirow{2}{*}{Methods} & \multicolumn{3}{c|}{$t=0$ to $t=1$} & \multicolumn{3}{c}{$t=1$ to $t=2$} \\ \cline{2-7}
& Old N/O & Overlapping & New N/O & Old N/O & Overlapping & New N/O \\ \hline
LwF~[16] & 0.00 & 20.13 & 31.74 & 0.00 & 10.30 & 27.50 \\
WA~[27] & 0.00 & 30.91 & 42.76 & 0.00 & 18.40 & 28.23 \\
iCaRL~[20] & 0.00 & 33.69 & 45.00 & 0.00 & 31.83 & 27.88 \\
IL2A~[29] & 0.00 & 34.45 & 41.67 & 0.00 & 36.78 & 29.19 \\
PASS~[30] & 0.00 & 34.61 & 46.25 & 0.00 & 28.16 & 28.65 \\
SSRE~[31] & 0.00 & 34.90 & \textbf{49.96} & 0.00 & 17.56 & 29.77 \\
CLVQA~[14] & 0.00 & 33.13 & 36.67 & 0.00 & 38.61 & 34.19 \\
CLiMB~[23] & 0.00 & 33.78 & 38.75 & 0.00 & 34.35 & 44.19 \\
Ours & 0.00 & \textbf{35.83} & 45.06 & 0.00 & \textbf{39.47} & \textbf{46.88} \\ \hline
\end{tabular}}
\end{table}

\end{document}